\documentclass[11pt]{article}

\usepackage[a4paper, margin=1in]{geometry}
\usepackage[utf8]{inputenc}
\usepackage[T1]{fontenc}
\usepackage[expansion=false]{microtype}
\usepackage{amsmath, amssymb}
\usepackage{graphicx}
\usepackage{booktabs}
\usepackage{xcolor}
\usepackage{titlesec}
\usepackage{enumitem}
\usepackage{url}
\usepackage[hidelinks]{hyperref}
\usepackage[numbers]{natbib}

\titlespacing*{\section}{0pt}{1.4em}{0.6em}
\titlespacing*{\subsection}{0pt}{1.1em}{0.4em}

\title{\textbf{Good Benchmarks}}

\author{Ivan Bercovich}

\date{July 2026}

\begin{document}
\maketitle

\begin{abstract}
\noindent
Good tasks are correct, solvable, verifiable, well-specified, and hard for interesting reasons. The best tasks describe a real problem an experienced practitioner would recognize, in language a practitioner would use, with tests that verify the outcome rather than the approach.
\end{abstract}

\begin{quote}
Benchmarks are where SOTA has to earn its name.
\end{quote}

This post is about designing good tasks. The examples come from Terminal Bench~\citep{tbench-original}, where contributors collectively spent thousands of hours developing and reviewing hundreds of them. The source material for this post came from conversations with \href{https://x.com/kjhennegen}{@kjhennegen}, \href{https://x.com/ryanmart3n}{@ryanmart3n}, \href{https://x.com/alexgshaw}{@alexgshaw}, \href{https://x.com/bla1990so}{@bla1990so}, \href{https://x.com/isegal}{@isegal}, \href{https://x.com/kexun_zhang}{@kexun\_zhang}, \href{https://x.com/StevenDillmann}{@StevenDillmann}, \href{https://x.com/never_settles_}{@never\_settles\_}, \mbox{\href{https://x.com/ekellbuch}{@ekellbuch}}, \href{https://x.com/LiBoxuan91538}{@LiBoxuan91538}, \href{https://x.com/AllenHa13152844}{@AllenHa13152844}, \href{https://x.com/lschmidt3}{@lschmidt3}, and many other TBench contributors.

Terminal Bench is a hard benchmark for agents, and we're trying to do realistic tasks. Evals used to ask if an AI understood something or could complete a specific, well-defined minitask. Agents are much harder to evaluate, both from the point of view of verification and the amount of compute it takes to know that an agent is working.

The best tasks describe a real problem an experienced practitioner would recognize, in language a practitioner would use, with tests that verify the outcome rather than the approach.

\section{What a good task is}

Good tasks are correct, solvable, verifiable, well-specified, and hard for interesting reasons. Every task must be:

\begin{enumerate}
    \item \textbf{Verifiable:} The task must be checkable with a program that is all-but-guaranteed to detect errors and all-but-guaranteed to accept correct solutions. It must be possible to make a reliable determination that separates correct solutions from incorrect ones. Reject tasks where grading is subjective (e.g., "make this game more fun"). Re-running the verifier hundreds of times should show no failures (e.g., short verifier timeouts, crashing).
    \item \textbf{Well-specified:} The problem description completely describes what a correct outcome looks like. Nothing ambiguous is left to guesswork. As a rough, intuitive guide, if two experts read the problem description, a well-specified problem should result in both of them writing verification algorithms so that anyone who passed one of the verifiers would also pass the other.
    \item \textbf{Solvable:} The problem must provide a solution, or a convincing argument that it is solvable. Someone who knows exactly what to do should be able to solve it within the time and resource limits of the task.
    \item \textbf{Difficult:} The task should be hard to solve. There are many reasons a problem could be difficult; we require it to be difficult for a "good reason."
    \item \textbf{Realistic and valuable:} There is a real scenario where a person is paid to do this, and the task mirrors how they actually do it. It is not enough to be interesting (e.g., a puzzle); a professional has to recognize it as their job.
    \item \textbf{Outcome-verified:} We are grading outcomes, not the process to get there.
    \item \textbf{Robust:} The task behaves the same across time and on any hardware that meets the stated resource requirements: no unpinned dependencies, no reliance on live external services.
\end{enumerate}

The ideal task can be described in two paragraphs and forces the agent to think before doing anything. A small program, under aggressive enough requirements, can be as difficult as a large one. The most elegant tasks have short, well-specified, self-explanatory instructions, without need for additional documentation. Think literate programming.

The goal isn't to make the task easy, but to make it unambiguous. What we want to avoid is a task that's hard only because it withholds a hint the agent needed; the difficulty should be intrinsic, not a missing fact.

\section{Start from real work}

Most of the first Terminal Bench tasks were things one of us had to do at work at some point in the past, converted to the appropriate format. Nobody's really looking for real-world challenges. Here's a real problem. Solve my freaking problem.

Terminal Bench was a grassroots effort. Most creators were individuals who were interested in the project, and most people made one or two tasks. The tasks were built by people with master's degrees and PhDs and by practicing engineers. Real tasks, very carefully curated. Some of them were very, very hard. There was a lot of peer review.

\section{What difficulty actually means}

The naive way to define a hard task is that if the models can't do it, then it's hard. But you can make something hard by making it tricky. You can make a deceptive puzzle with hints that throw you the wrong way. Relativity is hard. A tricky puzzle is just tricky.

This has been a huge issue: tasks that are trivial once well specified. The trouble is the mismatch between what we ask for and what we actually want; the gap is in what's implicit, the things you assume are obvious when assigning a task but fail to specify.

A lot of tasks start with a prompt that works, and then the author removes tokens until it fails. You basically start with a task that is solvable and progressively make it less solvable. It's almost like reverse backprop/GEPA to make the task hard. It doesn't really make it hard.

The hardest tasks tend to be hard the first time. What we want to avoid is finding a medium-hard task and going into the instruction to make it artificially hard by leaving some ambiguity there. The task tends to be hard because it is outside the jagged frontier~\citep{jagged-frontier}, not because you give it half the time or require it to be twice as fast.

So we come back to tasks where the description is straightforward and unambiguous, the verifier robust, but the answer hard.

There's a categorical difference between a task that fails because the expected output is structurally convoluted (a long instruction explaining how to form a complex JSON object) and one that fails for something intrinsically difficult. If an agent fails because it put a dollar sign in the amount field when you wanted a float, or used a top-level key when you wanted a bare array, you're measuring format compliance. That's not the capability a frontier benchmark sets out to test. If a task is hard for SOTA models, it shouldn't be because they can't spell "strawberry." A related problem is tasks that are too wide, asking for a lot of little deliverables instead of one concrete large problem.

Some tasks are the concatenation of many tasks. The probability of going off the rails is higher, but any one part may not be hard. Other times, coordination between multiple tasks is the problem you are going after. Then holding the process together and driving it to the finish line is the point.

Sometimes the agent near-misses. It gets 90 subgoals right and misses 5. If those 5 are the hard ones, could the task just focus on them? If the near-miss is a single quantitative threshold, the fix is a new conceptual challenge in the same domain, not a tighter threshold.

After a bug in the task is fixed, the task sometimes goes from very hard to very easy. That is a reason to look closer at what was making it hard in the first place.

\section{Focus on the outcome}

The best instructions are brief, state the final goal upfront, and give the agent latitude to deliver on the objective. Instructions that enumerate a long list of steps or procedures the agent must follow are often overspecified. Keep it clear, but not redundant. Don't tell the agent how it will be tested, but tell it what you expect the end result to be, where the instructions are sufficiently specified such that to meet them implies passing the tests.

Even when instructions are human-written, they're often too prescriptive. Authors tell the agent how to solve the problem instead of what the end state should be. It's too clerical, a specific set of steps, instead of a goal. Better to describe how the system should work and let the agent figure out the rest. You don't need to spell out how things might fail: assume an experienced engineer would understand what you mean, and expect the agent to bring the same understanding.

We treat every token as a chance to mistakenly add ambiguity, or to create a specification detail the tests might miss. If the task is to be unambiguous and perfectly verified, brevity is a KPI.

A useful heuristic is to be as broad as possible and as specific as necessary, because that's how humans actually use instructions. When an instruction is broad enough to allow several reasonable approaches, either the verifier has to accept all of them, or the instruction needs to be tighter. Otherwise agents fail for guessing the wrong convention. Tightening the instruction is usually the simpler fix, because verifiers are the hard part.

In practice, very few people write a full page of precise requirements by hand before handing an agent a task. When a task does arrive with one, it's usually AI-generated, with nobody auditing the specifics, so an instruction can be perfectly well specified and still wrong.

\section{Prove that the task is solvable}

The submitted solution should solve the problem the way an agent would. Hardcoding an answer that implies knowledge not self-evident in the instructions doesn't help: you’re making assumptions the instructions don't state, so the task is underspecified, and you don't realize it until you read the solution.

A proper oracle for a diagnostic task interrogates the system. It runs a series of commands to figure out what's wrong, and only then produces a solution. If the solution jumps straight to the answer with no exploratory steps, it may be making unfair assumptions: the author knows the ground truth, so the solution isn't investigating the issue, just writing down the answer to a known one.

The deeper test is whether the author can give a complete, self-contained path from the instructions to the verified answer: every step, at the level of detail acceptable in their field, ending in the answer the verifier accepts. They had to do this work to write the verifier in the first place; if they can't produce it now, they probably don't understand the task. And if you can't confirm the solution is correct, you can't trust the verifier, especially when it checks exact values, where a basically-correct solution with a single sign error is still broken.

A lot of these underspecified tasks are the result of someone building collaboratively with an AI and then removing pieces of the solution, with no evidence that the removed piece can be reconstructed from the rest.

\section{Build a controlled environment}

The task should behave the same across time and on any hardware that meets the stated resource requirements. No unpinned dependencies, no reliance on live external services.

The environment has to contain everything the agent needs to solve the task, and the oracle has to work from that same information. It cannot rely on privileged information about the task. This is common with synthetic tasks, where information is scrambled or corrupted and the agent has to recover it.

Then ask whether the environment is hackable~\citep{terminal-wrench-2026,krakovna2020specification,metr2025hacking}. Can you get the reward without doing the task? A trajectory is one instance of a hack, but you still have to explain what makes the environment hackable. Is it the test? Is it something about the way the environment is set up? Why is the agent able to produce the hack, and how would you mitigate it?

The weakness may belong to one task, or it may be something fundamental about the harness that affects every task built on it. After fixing it, run the normal solution again. A fixer can make the task impossible too.

\section{Make the verifier mean something}

Tests should verify outcomes, not implementations. Why check that Pandas is installed, when the task never asked for it? Evaluating source code by string comparison is brittle, and you should test on examples beyond the one you handed the agent.

Tests that check for specific libraries instead of correct output, or that are so tightly coupled to the oracle solution that any other correct approach fails, are common. The root cause is often a verifier co-designed with the author's own solution: built backwards from one answer rather than derived from the instruction, so it isn't flexible enough to accept other valid ones.

When you start with a problem, it tends to be straightforward. The verifier is tricky because there may be multiple ways to arrive at the solution. You keep running agents and see that they're trying this and they're trying that. Maybe they're all failing, but you have to be open to appreciating that there are other ways to arrive at the solution. The verifier has to accept those other correct paths. To me, that's a discovery process.

This is especially easy to miss with synthetic data. When you generate the data from a hypothesis, the verifier can quietly encode an assumption only you hold: it works for your formulation and no other. It's like inventing a simple-looking cipher that's actually impossible, because only you know the answer you built it around.

One way to check whether a verifier is fair rather than co-designed with one answer: temporarily over-specify the instruction (name the method, add hints) and confirm a well-guided agent can produce an answer the verifier accepts. If it can, the difficulty was in choosing the approach, not in an arbitrary verifier, and the verifier is fair conditional on the method. If a well-guided agent still fails, look harder at the verifier.

What about LLM-as-a-judge? An LLM judge will find some issues, but not everything. On genuinely hard tasks, we've seen a judge say the instruction was insufficient because a test assumed that a person was shorter than three meters. That is a perfectly reasonable assumption. As of right now, you still need a human to look at the trajectory and see what the mistake is.

Grade outcomes, not process. Solutions should be deterministic, but the problem can still be dynamic. People tend to fork on one question: should the AI give you the answer, or give you a piece of software that produces the answer? The latter tends to be more verifiable, more testable. A task can't say "use emacs to edit a file"; vim is allowed. The one place a process constraint is justified is as an anti-cheat measure.

\section{Run the task and watch it fail}

When we get stuck debugging, we run the task with the oracle, interact with the container, run the tests, and see what happens. If the failure came from an agent solution, convert the agent log into an alternative solution and step through it.

Look at the logs of agents trying to solve the task, the failures especially, and work out why. Did they fail because it's hard or because it's unfair? Because the instructions were insufficient, or the tests too aggressive? Or because they genuinely didn't know what to do?

You have to convince yourself the failures aren't from a deficiency in the task. You might be skeptical that a SOTA model is failing something that looks easy and one-shot, so run trials in batches of 5, dig in with your harness's debug tooling, and see what the failures have in common. These are the insights you get from watching trajectories.

And check crux alignment: does what actually broke match what you said would be hard? If the stated crux is one thing but every failing trial breaks on something else, the task is hard for the wrong reasons, and either the instruction, the environment, or the difficulty claim has to change.

None of these alone is disqualifying; each is a reason to look closer.

\begin{itemize}
    \item Passing rate swings a lot on a tiny change: a pre-installed library, a package version, a small hint in the instructions.
    \item Multiple agent failures converge on the same wrong answer. Sometimes this is fine, but it's usually fishy.
    \item Deep review keeps returning critical and major concerns even after two or three iterations. Submissions where feedback doesn't converge tend to be problematic.
    \item Instructions have a large surface area: asking the agent to do a lot of things, look at a lot of files, and so on.
    \item The solution is hardcoded: rather than the series of steps a person or agent would take, it just states the answer, for example an encryption task that knows the key.
    \item The solution has privileged information about the task.
    \item The task is too synthetic, with no clear relation to a real-world scenario.
    \item Check any outliers: high variability between runs in passing rate, cost, and so on.
\end{itemize}

\section{Keep authorship and review human}

Someone asks an LLM to write their task instructions and submits whatever comes out. It's immediately recognizable: the tone is wrong, it's verbose, it's over-structured, and it reads like it was written to maximize the probability the agent succeeds.

Great instructions are written by hand, or heavily edited from what an LLM suggests. They aren’t a forced prompt that uses emphasis and repetition to coerce the attention heads into listening. Direct and to the point. Specific and sufficient, but not redundant and attention-grabbing.

AI is really good at reviewing tasks. But if the author is also using AI to handle the feedback, it becomes an AI dialogue and the tasks become worse. The question is how you extract as much intelligence as possible from the AI without the author becoming brain dead. If you agree with everything the AI tells you, you end up hill climbing toward a hard task that is too artificial.

A lot of tasks end up gradient descending either some AI review or some reviewer. The task wasn't good to begin with. Then it tries to get all the checkboxes and becomes a bad artificial task.

There are two kinds of feedback you can give on a task. One is benchmark correctness: is this a valid task, is the metadata complete, is the instruction in shape. That makes a task cleaner and tighter. The other is deeper correctness, and it takes a domain expert. What we got wrong in an earlier generation of Terminal Bench was assigning random reviewers. They were good; they'd made tasks themselves, so they could point out what to clean up. But that's not the priority. Cleaning up a bad idea doesn't make it a good task; if the underlying task is broken, tidying it doesn't help.

So from the earliest stages, feedback should come from someone who can actually solve the task. That's the bar: a reviewer who could solve it themselves, and who has made a task before.

Being a domain expert and being able to build a good task are different skills. Writing a task well means thinking like an agent-evaluator, anticipating the many ways an agent could legitimately reach the answer and where it could reward hack, which is not the headspace most domain experts live in. A domain expert is great at finding problems but can rarely certify that a task is correct, because certifying it takes an understanding of agent evaluation, not just the domain. It cuts the other way too: highly technical, specialized-sounding work can fool a non-expert reviewer into thinking a task is good. There's good complexity and bad complexity, and telling them apart takes a domain expert.

One model that works is having a reviewer shepherd the task end to end, a little like pair programming. Someone submits a task; you discuss it with them, a fifty-minute call or asynchronously, making sure they've read the guidelines and thought about them. Later you both review it again, with more feedback. By the end you actually know the task: its structure, which files matter. Some tasks have hundreds of files, many of them background, so it matters to know what you're looking at. Ideally the same person works with the author from idea to merge.

Domain-specific reviews tend to snag in the same loop: the reviewer thinks the task is underspecified, the author says this is standard in our field. You need a standard of evidence to point to, the way Wikipedia has policies: you can't just assert that it's standard in your field, you have to cite an authoritative source. That kills most of the back-and-forth. It's also worth specifying the disputed piece of information and seeing what happens: if the task drops from very hard to very easy, it wasn't testing agentic behavior, it was testing a one-sentence fact the agent didn't know, and the task is weak either way.

\section{Why this matters}

AI is a deeply empirical field, and without hands-on experience it's hard to build the right intuition. When we started soliciting tasks for Terminal Bench 3, we rejected the first several proposals; then, like the 4-minute mile, once one good one came through, many followed. Benchmarks are where you build a feel for what the next months and years might hold. Anyone can get access to AI; most people have no real sense of where we are on the capability curve.

It's a little crazy that major labs outsource semi-artificial tasks while people solve real problems every day. People are doing things with LLMs that could be tasks without realizing it. Every week you spend a few hours solving a problem—that's a task. The best tasks take real time. Some have taken months of back-and-forth and well over forty hours of work between a reviewer and an author, both part-time experts. It’s very hard to build a SOTA benchmark for volume.

In the end it's mostly about authors doing good work. Reviewers can tighten a good task or kill a bad one, but you can't handhold someone into a good task. Most benchmarks would be better off with fewer, better tasks.

Quality over quantity.


\begin{thebibliography}{99}

\bibitem[Bercovich et al.(2026)]{terminal-wrench-2026}
I.~Bercovich, I.~Segal, K.~Zhang, S.~Saxena, A.~Raghunathan, and Z.~Zhong.
\newblock Terminal {W}rench: A dataset of 331 reward-hackable environments and 3,632 exploit trajectories.
\newblock \emph{arXiv preprint arXiv:2604.17596}, 2026.

\bibitem[Dell'Acqua et al.(2023)]{jagged-frontier}
F.~Dell'Acqua, E.~McFowland~III, E.~R. Mollick, H.~Lifshitz-Assaf, K.~Kellogg, S.~Rajendran, L.~Krayer, F.~Candelon, and K.~R. Lakhani.
\newblock Navigating the jagged technological frontier: Field experimental evidence of the effects of {AI} on knowledge worker productivity and quality.
\newblock Harvard Business School Working Paper No.~24-013, 2023.

\bibitem[Krakovna et al.(2020)]{krakovna2020specification}
V.~Krakovna, J.~Uesato, V.~Mikulik, M.~Rahtz, T.~Everitt, R.~Kumar, Z.~Kenton, J.~Leike, and S.~Legg.
\newblock Specification gaming: the flip side of {AI} ingenuity.
\newblock DeepMind Blog, 2020.

\bibitem[Merrill et al.(2026)]{tbench-original}
M.~A. Merrill et al.
\newblock Terminal-{B}ench: Benchmarking agents on hard, realistic tasks in command line interfaces.
\newblock In \emph{Proceedings of the International Conference on Learning Representations (ICLR)}, 2026.
\newblock arXiv:2601.11868.

\bibitem[Von Arx et al.(2025)]{metr2025hacking}
S.~Von~Arx, L.~Chan, and E.~Barnes.
\newblock Recent frontier models are reward hacking.
\newblock METR Blog, 2025.
\newblock \url{https://metr.org/blog/2025-06-05-recent-reward-hacking/}.

\end{thebibliography}
\end{document}